# Domain Wall Leaky Integrate-and-Fire Neurons with Shape-Based Configurable Activation Functions

Wesley H. Brigner[1], Naimul Hassan[1], Xuan Hu[1], Christopher H. Bennett[2], Felipe Garcia-Sanchez[3], Can Cui[4], Alvaro Velasquez[5], Matthew J. Marinella[2], Jean Anne C. Incorvia[4], Joseph S. Friedman[1]

[1]Electrical and Computer Engineering, University of Texas at Dallas, Richardson, TX, United States
[2]Sandia National Laboratories, Albuquerque, NM, United States
[3]Department of Applied Physics, University of Salamanca, Salamanca, Spain
[4]Electrical and Computer Engineering, University of Texas at Austin, Austin, TX, United States
[5]Information Directorate, Air Force Research Laboratory, Rome, NY, United States

## ABSTRACT

Complementary metal oxide semiconductor (CMOS) devices display volatile characteristics, and are not well suited for analog applications such as neuromorphic computing. Spintronic devices, on the other hand, exhibit both non-volatile and analog features, which are well-suited to neuromorphic computing. Consequently, these novel devices are at the forefront of beyond-CMOS artificial intelligence applications. However, a large quantity of these artificial neuromorphic devices still require the use of CMOS, which decreases the efficiency of the system. To resolve this, we have previously proposed a number of artificial neurons and synapses that do not require CMOS for operation. Although these devices are a significant improvement over previous renditions, their ability to enable neural network learning and recognition is limited by their intrinsic activation functions. This work proposes modifications to these spintronic neurons that enable configuration of the activation functions through control of the shape of a magnetic domain wall track. Linear and sigmoidal activation functions are demonstrated in this work, which can be extended through a similar approach to enable a wide variety of activation functions.

## 1 Introduction

According to neuroscientists, the human brain consists of neurons and synapses. Neurons integrate input electrical pulse trains through their axons. When enough input pulses have been received, these cells release output pulses from their somas, through their dendrites, and into the axons of other neurons. Synapses bridge the gaps between two neurons.

Likewise, artificial neuromorphic systems consist of elementary neuron and synapse components. They can be implemented using software run on standard von Neumann computers [1],[2], but such a method is highly inefficient due to the fact that conventional mathematical operations are performed sequentially rather than in parallel, as to would biological neuronal and synaptic functions. Furthermore, complementary metal-oxide semiconductor (CMOS) technology does not natively provide the required neuronal or synaptic functionality – these behaviors must be emulated with large circuit overhead. The efficiency can be improved by designing specialized CMOS architectures specifically for neuromorphic applications [3],[4]; however, even though this will significantly reduce the device count, and therefore the power consumption, CMOS devices are still not ideal for these applications due to their volatile and digital nature.

The non-volatility and analog nature of spintronics is particularly attractive for neuromorphic computing, and several beyond-CMOS spintronic synapses and neurons have been proposed to improve the efficiency. Whereas synapses only require non-volatility and a variable resistance, the popular leaky integrate-and-fire (LIF) neuron model requires three primary functionalities: leaking, integrating, and firing. Therefore, while much progress has been made on beyond-CMOS synapses [5]-[8], significant challenges remain to emulate the complex behavior of neurons. Advancing on previous work that required CMOS within the network [8], we have previously proposed spintronic LIF neurons capable of intrinsically providing the three necessary functionalities within a purely spintronic system [9]-[13].

However, these neurons have limited capability to mimic the various activation functions commonly used in neural networks. Activation functions are commonly used in neural networks to improve the decision making process and, by extension, the learning characteristics of the networks. A variety of activation functions are utilized for neural networks, including the rectified linear unit (ReLU) and squashing/sigmoidal activation functions.

In this paper, we propose spintronic neurons that provide these activation functions by modifying the shape of one of our previously-proposed shape-based magnetic domain wall neuron [11]. By appropriately configuring the shape of this device, it is possible to implement a wide range of distinct activation functions. This permits the design of neural networks that leverage nearly any arbitrary activation function, thereby increasing the efficiency of the spintronic neuromorphic networks

Section 2 provides a brief background into the field of neuromorphic computing, including the shape-based neuron. Section 3 discusses the realization of two particular activation functions, while conclusions are provided in section 4.

## 2 Background

Neuromorphic computing is generally realized with neuron layers connected through crossbar synapse arrays. In order to realize the LIF neuron functionality, we have previously proposed the use of artificial spintronic neurons as described in this section. This section also overviews the activation functions of interest for neuromorphic computing.

### 2.1 Crossbar Array

Crossbar arrays typically consist of horizontal input wires (word lines) and vertical output wires (bit lines). Input neurons are placed at the inputs to the word lines, and output neurons are placed at the outputs of the bit lines. Synapses, on the other hand, are placed at the intersections of the word and bit lines. Therefore, an MxN crossbar array will consist of M+N neurons and M*N synapses [14]-[16].

### 2.2 Leaky Integrate-and-Fire Neuron

In order to accurately mimic biological neurons for neuromorphic computing, artificial LIF neurons implement three primary functionalities: leaking, integrating, and firing. When integrating, these neurons accept and store energy from input energy pulses. When no input pulse is provided, the stored energy gradually dissipates. Finally, once sufficient stored energy has been integrated, the neuron releases this energy as an output pulse of its own.

### 2.3 Domain Wall Magnetic-Tunnel-Junction

Magnetic tunnel junctions consist of two ferromagnetic layers – a "free" layer capable of changing states and a "fixed" or "pinned" layer whose magnetization is stable. When the two layers are magnetized parallel to each other, the device exhibits a low-resistance state (LRS); when they are magnetized anti-parallel to each other, the device exhibits a high-resistance state (HRS). Domain wall-magnetic tunnel junctions (DW-MTJs) are similar, but the free layer is extended and contains two anti-parallel magnetic domains bounded by a domain wall [17],[18]. The DW can be moved with a spin-injected current through a heavy metal beneath the DW track, and the device changes resistance states when the DW shifts underneath the MTJ. Recent experimental demonstrations suggest that this spin-injected current results in the extremely energy efficient spin-orbit-torque style within the output terminal [19].

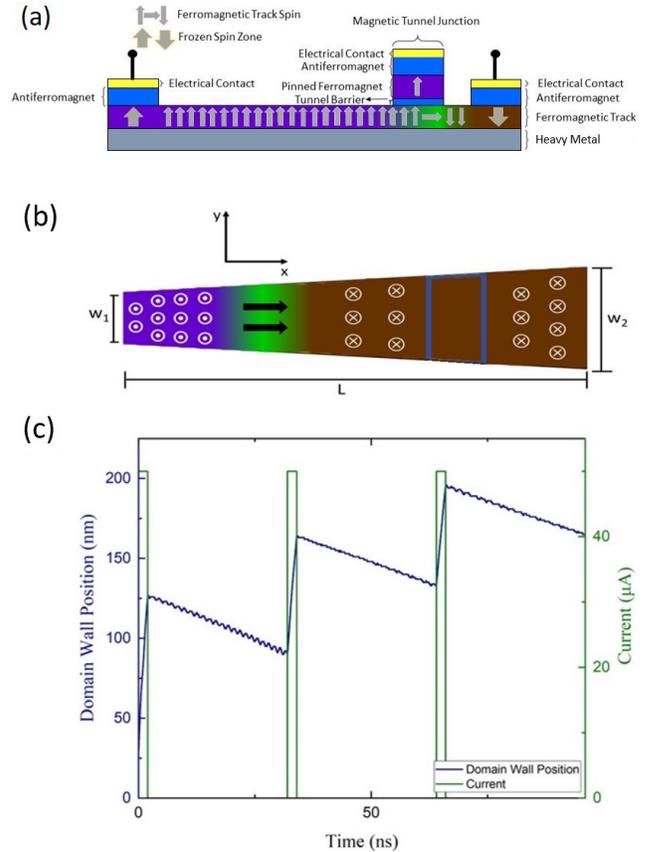

Figure 1: (a) Side view of a DW-MTJ neuron. (b) Top-view of the neuron with shape-based DW drift. (c) Combined integrating and leaking characteristics of the neuron with shape-based DW drift.

### 2.4 DW-MTJ LIF Neuron

This DW-MTJ device can be used as an LIF as shown in Figure 1(a), with the neuron energy represented by the position of the DW within the track [9]. Integration is accomplished by applying current through the heavy metal, and firing occurs when the DW passes underneath the MTJ, thereby switching the current across the tunnel barrier.

In order to induce leaking that shifts the DW in the direction opposite the SOT current, an energy landscape must be produced that causes the DW to exist in a lower energy state at one end of the device than the other. While this can be achieved by providing current through the DW track in the direction opposite the integration, this approach is undesirable due to the increased power consumption, Ohmic heating, and additional control circuitry [8]. It is preferable, therefore, for the leaking to be passive, as in [9]-[12].

The shape-based leaking of [11] is particularly attractive for enabling useful activation functions. With this method, the DW

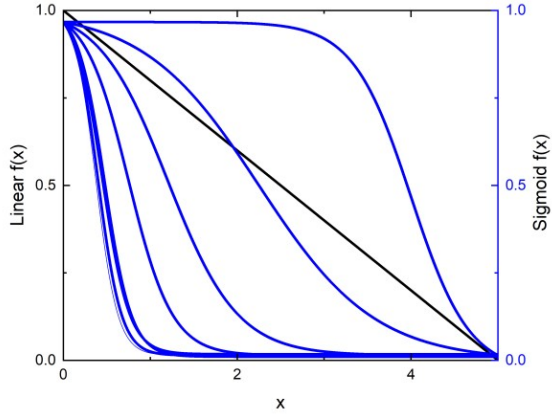

**Figure 2:** Generalized linear (black) and sigmoidal (blue) activation functions. The sigmoidal activation functions are shown with various switching speeds.

track width is varied from one end of the track to the other, as shown in Figure 1(b). DWs typically exist in lower energy states in wider tracks than in narrower tracks. Consequently, this variation of the DW track width creates an energy landscape more favorable to the DW existing on the left side of the track than on the right side, causing the DW to shift from right to left.

The integrating and leaking characteristics observed in mumax3 micromagnetic simulations are illustrated in Figure 1(c) [11]. The magnetic parameters are listed in Table I [20]. These micromagnetic parameters are used for the entirety of this work, including the linear and squashing neurons of section 3. Throughout this work, COMSOL has been used to create a current map for non-rectangular DW-MTJ neuron structures.

## 2.5 Activation Functions

Activation functions allow a neuron to provide the network with significantly improved learning characteristics during training, and significantly improved performance during operation. In fact, it has been shown that particular activation functions, such as the linear or sigmoidal activation functions shown in Figure 2, can reduce the error exhibited by a neural network by up to two orders of magnitude when the network is applied to certain data sets [21].

| Parameter | Value |
|---|---|
| Exchange Stiffness (Aex) [J/m] | $13*10^{-12}$ |
| Landau-Lifshitz Damping Constant (alpha) [dimensionless] | 0.05 |
| Non-Adiabaticity of STT (xi) [dimensionless] | 0.05 |
| Saturation Magnetization (Msat) [A/m] | $7.958*10^5$ |
| First Order Uniaxial Anisotropy Constant (ku1) [J/m$^3$] | $5*10^5$ |

**Table I:** Material parameters used in the micromagnetic simulations. The parameters shown here correspond to those exhibited by CoFeB.

The ReLU activation function maps positive inputs to outputs in a linear fashion. In addition, as the physical implementation of our ReLU function has a natural bound for maximum and minimum activation, and can be easily read at several analog levels, it is especially amenable to physical implementation of quantized neural networks [22],[23]. On the other hand, the sigmoid function (also referred to as the squashing function), maps the input to a monotonically decreasing output, with the highest rate of change at the center of the function. Sigmoid functions, and a mathematically adjusted version of this activation function known as the hyperbolic tangent (*tanh()* ) function, are more heavily used in recurrent neural network designs where they provide expressive capabilities and stability [24],[25]

## 3 DW-MTJ Neurons with Configurable Activation Functions

In order to improve the biomimetic capabilities of our neurons, it is important for them to implement a variety of activation functions, including the linear and squashing activation functions. To do so, modifications to the neurons are required. The neuron from [11] is particularly well suited to implementing these functions due to the simplicity of the necessary changes. Although this section only demonstrates the linear and squashing activation functions, it is clear that this approach can be extended to a wide variety of activation functions using similar modifications.

It is important to note that while the leaking behavior is dependent solely on the neuron structure, the integration behavior is also dependent on the applied input current magnitude. Therefore, in order to ensure that the desired activation functions are always significantly impacting the neuron behavior, these activation functions are implemented in terms of the leaking rather than the integration. For any given leaking activation function, a wide range of integration activation functions can be achieved by varying the current magnitude applied to the DW-MTJ neuron.

### 3.1 Neuron with Linear Activation Function

In the trapezoidal DW-MTJ device of Figure 1(b), the DW accelerates as it nears the narrow end of the track. To realize a linear activation function without this acceleration, it is necessary to alter the shape of the DW track to decrease the leaking force in narrower regions of the track.

Linear leaking similar to the black line in Figure 2 can be accomplished simply by introducing a slight exponential variation in the width of the DW track, as shown in Figure 3(a). In general, as the value of *b* increases, the linearity of the device's leaking increases, calculated as the inverse of the root mean squared error

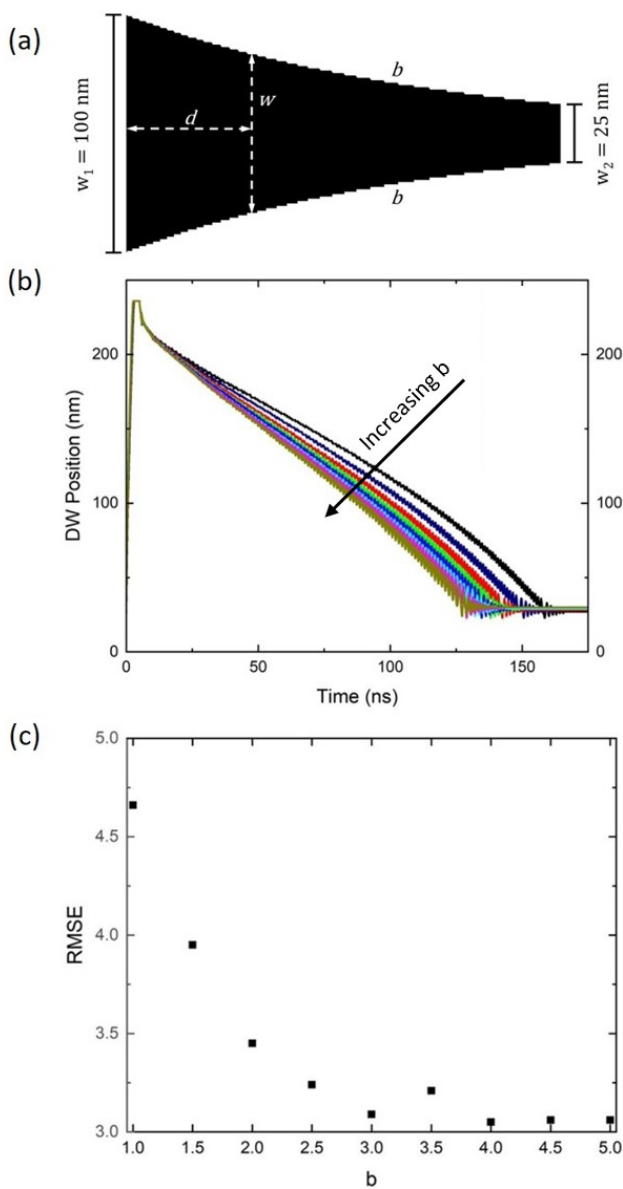

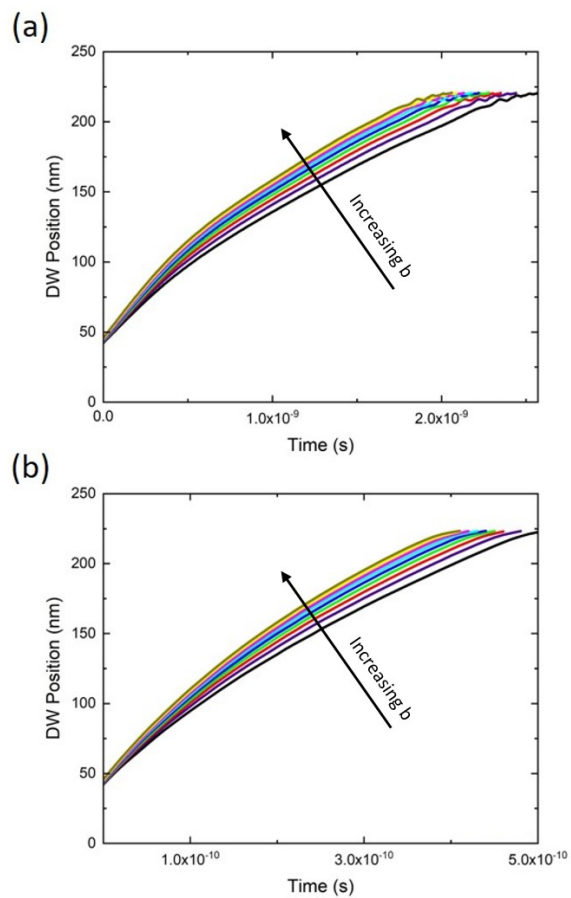

Figure 4: (a) Integration of the linear neuron for an input current of 0.1 mA. As with Figure 4(b), *b* ranges from 1 to 5 with an increment of 0.5. (b) Integration of the linear neuron for an input current of 0.5 mA.

It is also important to analyze the response of the DW-MTJ neurons to various input currents. When an input current of 0.1 mA is applied to the neuron, as in Figure 4(a), the DW's integration speed increases in proportion to *b*. As the DW nears the end of its range of motion, it also begins to exhibit slight oscillatory behavior due to interactions with the fixed region at the edge of the neuron. When the input current is increased to 0.5 mA, as shown in Figure 4(b), the integration speed also increases, as would be expected. Additionally, the integration speed maintains its positive correlation with the value of *b*, but the higher current prevents the previously observed oscillations of the DW at the edges of its range of motion.

### 3.2 Neuron with Squashing Activation Function

In order to implement sigmoidal leaking, the leaking force must not only be minimized at the narrow end of the track, but also at the wide end of the track. Therefore, the neuron's shape gradient can of a linear regression performed on the leaking curve. The leaking speed also increases as *b* increases. However, once *b* reaches a certain point, further increases cease to produce an increase in linearity, although they continue to increase leaking speed.

Figure 3: (a) Top view of the linear neuron, displaying the slight exponential curvature of the sides of the track. This curvature is of the form $w \propto b^{-d}$, where *b* represents the curvature of the sides, *d* is the distance from the wide end of the track, and *w* is the width of the device at distance *d*. *b* ranges from 1 to 5 at intervals of 0.5. When *b* = 1, the sides are straight, and the track is identical to the one shown in Figure 1(b). (b) Leaking characteristics of the linear neuron for various values of *b*, including *b* = 1. (c) Root mean squared error (RMSE) of the neuron's leaking characteristics from a linear function.

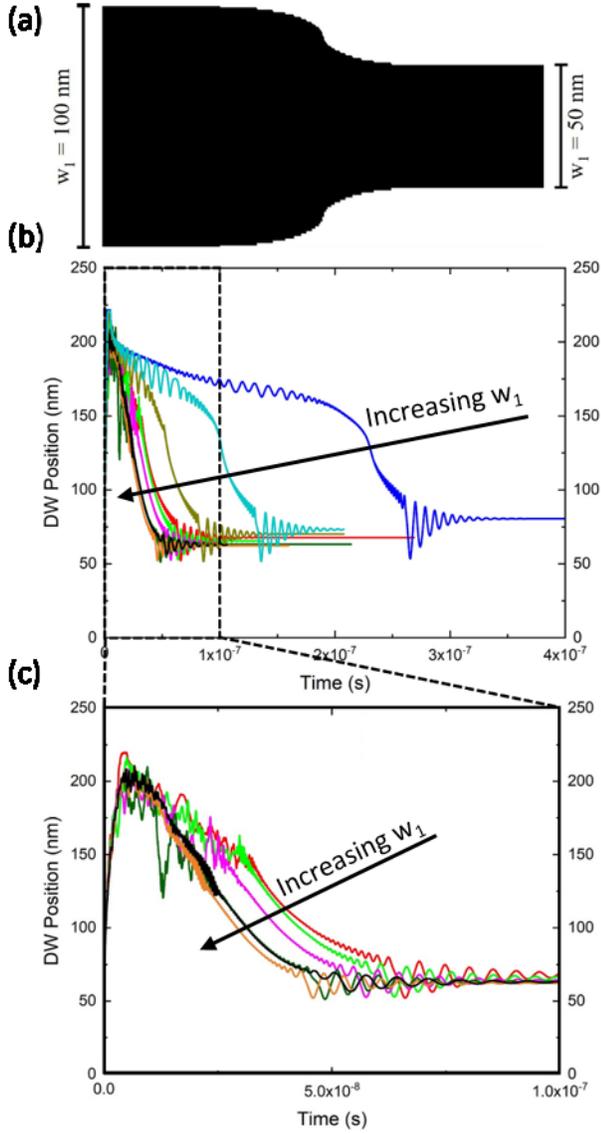

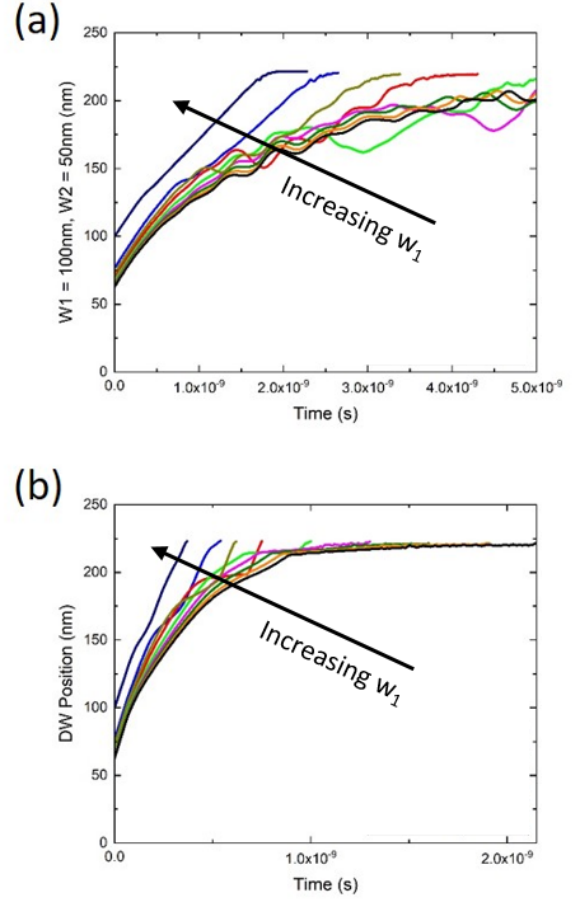

**Figure 5:** (a) Top view of the squashing neuron, displaying the sharp constriction of the DW track centered in the middle of the track. (b) Leaking characteristics of the squashing neuron for $w_1$ increasing from 100 nm to 400 nm, with an increment of 50 nm. (c) Leaking characteristics of the squashing neuron for $w_1$ increasing from 150 nm to 400 nm, into the time range 0 s to 100 ns.

only exist within a narrow range halfway between the narrow and wide ends of the DW track, as illustrated in Figure 5(a).

With $w_1$ = 100 nm as in Figure 5(b), the neuron exhibits a sigmoidal leaking characteristic similar to the blue curves in Figure 2. As the width of the wide end of the track increases, the DW leaking motion becomes faster, with an effect similar to that of $b$ on the leaking speed of a linear neuron. By zooming in on the leaking characteristics for these larger values of $w_1$ in Figure 5(c), it can be observed that the neurons still implement squashing functions.

**Figure 6:** (a) Integration of the squashing neuron with an input current of 0.1 mA. (b) Integration of the squashing neuron with an input of 0.5 mA. As $w_1$ increases, the steady state fully-leaked initial position approaches the end of the track.

As with the previously discussed linear neuron, it is important to analyze the integration characteristics of the squashing neurons for various input currents. With an input current of 0.1 mA, illustrated in Figure 6(a), the DW integration speed is inversely proportional to $w_1$, partly due to an increased leaking force and partly due to a decreased current density. Additionally, as the width increases, the integration becomes non-monotonic due to instability of the wide DWs. When the input current is increased to 0.5 mA as in Figure 6(b), not only does the DW integration speed increase significantly, the integration speed remains inversely related to $w_1$. Additionally, with increased current, the integration becomes monotonic even with large widths.

## 4 Conclusion

Shape-based DW drift enables configurable DW-MTJ neuron leaking that enables the realization of diverse activation functions for efficient learning and recognition in spintronic neuromorphic computing systems. In this work, we have demonstrated linear and squashing activation functions through specific configurations of the shape of the DW tracks. By extension of this concept, further activation functions commonly used in the field of neuromorphic can also be realized. This represents a significant advancement over previous spintronic neurons that will enable drastically improved learning characteristics of spintronic neural networks, and paves the way towards state-of-the-art neural networks being complemented with DW-MTJ synapses and neurons for on-chip learning

## ACKNOWLEDGMENTS

The authors thank E. Laws, J. McConnell, N. Nazir, L. Philoon, and C. Simmons for technical support, and the Texas Advanced Computing Center at The University of Texas at Austin for providing computational resources. This research is sponsored in part by the National Science Foundation under CCF awards 1910800 and 1910997, the National Science Foundation Graduate Research Fellowship under Grant No. 1746053, the Eugene McDermott Graduate Fellowship Award No. 202001, and the Texas Analog Center of Excellence Graduate Fellowship. Sandia National Laboratories is a multimission laboratory managed and operated by NTESS, LLC, a wholly owned subsidiary of Honeywell International Inc., for the U.S. Department of Energy's National Nuclear Security Administration under contract DE-NA0003525. This paper describes objective technical results and analysis. Any subjective views or opinions that might be expressed in the paper do not necessarily represent the views of the U.S. Department of Energy or the United States Government.

## DATA AVAILABILITY STATEMENT

The data that support the findings of this study are available from the corresponding author upon reasonable request.